# Tacit knowledge mining algorithm based on linguistic truth-valued concept lattice

Li Yang[1], Yuhui Wang[2]

(1. *School of Mathematics and Information Science, North China University of Water Resources and Electric Power, Zhengzhou 450011, China*

2. *College of Traffic & Transportation, Southwest Jiaotong University, Chengdu 610031, China*)

**Abstract**

This paper is the continuation of our research work about linguistic truth-valued concept lattice. In order to provide a mathematical tool for mining tacit knowledge, we establish a concrete model of 6-ary linguistic truth-valued concept lattice and introduce a mining algorithm through the structure consistency. Specifically, we utilize the attributes to depict knowledge, propose the 6-ary linguistic truth-valued attribute extended context and congener context to characterize tacit knowledge, and research the necessary and sufficient conditions of forming tacit knowledge. We respectively give the algorithms of generating the linguistic truth-valued congener context and constructing the linguistic truth-valued concept lattice.

## 1. Introduction

In our living environment, tacit konwledge (as opposed to formal or explicit knowledge) is everywhere, which is usually difficult to transfer to another person by means of writing it down or verbalising it. With tacit knowledge, people are often aware of the knowledge they possess or how it can be valuable to others and effective transfer of tacit knowledge generally requires extensive personal contact and trust. According to Parsaye [1], there are three major approaches to capture the tacit knowledge from groups and individuals: (i) Interviewing experts: structured interviewing of experts in a particular subject is the most commonly used technique to capture pertinent, tacit knowledge; (ii) Learning by being told: this can be done by interviewing or by task analysis. And task analysis is the process of determining the actual task or policy by breaking it down and analyzing what needs to be done to complete the task; (iii) Learning by observation: this can be done by presenting the expert with a sample problem, scenario, or case study and then observing the process used to solve the problem.

  However, the above methods are more complicated to use and lack of rigorous mathematical foundation. In this paper, we select the concept lattice model to capture tacit knowledge through depicting and analyzing explicit knowledge. Concept lattice (also called formal concept analysis FCA) was proposed by Wille [2-4] in 1982，and its ideological core is constructing the binary relation between objects and attributes based on bivalent logic. Facing the massive fuzzy information existed in reality, fuzzy concept lattice has appeared [5-8], which is used to describe the fuzzy relation between objects and attributes. As a conceptual clustering method, concept lattices have been proved to benefit machine learning, information retrieval and knowledge discovery, etc.

As we all know, natural language is one of the most commonly direct ways that people express ideas and transmit information, especially for tacit konwledge. Linguistic truth-valued concept lattice researched in this paper, different from the classical concept lattice and the fuzzy concept lattice, is a new mathematical model for dealing with linguistic information, which is constructed based on the linguistic truth-valued lattice implication algebra. Its key point of this model is that its values range is not general interval [0,1] but a complete lattice structure, on which incomparability linguistic information can be considered very well. This selection of lattice implication algebra has two advantages in contrast to general structure: firstly, the computational process of linguistic truth-valued concepts is closed induced by the operations upon the lattice implication algebra; secondly, the complex derivation is avoided in setting up the Galois connection of linguistic truth-valued concept lattice. A large amount of research on linguistic truth-valued concept lattice may be referred to [9-13]. While for lattice implication algebra, Xu [14-16] proposed this concept by combining lattice and implication algebra in 1980's in order to depict uncertainty information more factually.

Based on these analyses, this paper establishes a simplified linguistic truth-valued concept lattice and puts forward tacit knowledge mining algorithm based on it. Specifically, we firstly utlizes a six-element linguistic truth-valued lattice implication algebra to depict natural language and define the relevant operations among them; secondly establishes the simplified linguistic truth-valued concept lattice based on this algebraic structure and express knowledge by linguistic truth-valued context; finally captures tacit konwledge through mining the relations among explicit knowledge and gives the relevant mining algorithm. This algorithm not only provides us a relatively simple method but also can be used to improve the reduction ability of linguistic truth-valued concept lattice from another point of view [17-19].

In Section 2, we give an overview of classical concept lattice and lattice implication algebra. In Section 3, the related works of linguistic truth-valued concept lattice are briefly summarized. Successively, the definitions of linguistic truth-valued attribute extended context and congener context are proposed in Section 4, where we show the relevant attribute extended judgment theorems based on congener context and give the generation algorithms of attribute extended context and linguistic truth-valued concepts, respectively. Concluding remarks are presented in Section 5.

**2. Concept lattice and lattice implication algebra**

In this section, we review briefly the classical concept lattices and lattice implication algebra and they are the foundations of constructing the linguistic truth-valued concept lattice.

**Definition 2.1** ( [2] ) A partial ordered set (poset) is a set in which a binary relation $\leq$ is defined, which satisfies the following conditions: for any $x, y, z$,

(1). $x \leq x$, for any $x$ (Reflexive),

(2). $x \leq y$ and $y \leq x$ implies $x = y$ ( Antisymmetry ),

(3). $x \leq y$ and $y \leq z$ implies $x \leq z$ ( Transitivity ).

**Definition 2.2** ( [2] ) Let $L$ be an arbitrary set, and let there be given two binary operations on $L$, denoted by $\wedge$ and $\vee$. Then the structure $(L, \wedge, \vee)$ is an algebraic structure with two binary operations. We call the structure $(L, \wedge, \vee)$ a lattice provided that it satisfies the following properties:

(1). For any $x, y, z \in L$, $x \wedge (y \wedge z) = (x \wedge y) \wedge z$ and $x \vee (y \vee z) = (x \vee y) \vee z$

(2). For any $x, y \in L$, $x \wedge y = y \wedge x$ and $x \vee y = y \vee x$.

(3). For any $x \in L$, $x \wedge x = x$ and $x \vee x = x$.

(4). For any $x, y \in L$, $x \wedge (x \vee y) = x$ and $x \vee (x \wedge y) = x$.

**Definition 2.3** ( [4] ) The formal context of classical concept lattice is defined as a set structure $(G, M, I)$ consisting of sets $G$ and $M$ and a binary relation $I \subseteq G \times M$. The elements of $G$ and $M$ are called objects and attributes, respectively, and the relationship $gIm$ is read: the object $g$ has the attribute $m$. For a set of objects $A \subseteq G$, $A^*$ is defined as the set of features shared by all the objects in $A$, that is, $A^* = \{m \in M \mid gIm \ \forall g \in A \}$. Similarly, for $B \subseteq M$, $B^*$ is defined as the set of objects that posses all the features in $B$, that is, $B^* = \{g \in G \mid gIm \ \forall m \in B \}$.

**Definition 2.4** ( [4] ) A formal concept of the context $(G, M, I)$ is defined as a pair $(A, B)$ with $A \subseteq G$, $B \subseteq M$ and $A^* = B$, $B^* = A$. The set $A$ is called the extent and $B$ the intent of the concept $(A, B)$.

**Definition 2.5** ( [14] ) Let $(L, \wedge, \vee, O, I)$ be a bounded lattice with an order-reversing involution $'$, $I$ and $O$ the greatest and the smallest element of $L$ respectively, and $\rightarrow: L \times L \rightarrow L$ be a mapping. If the following conditions hold for any $x, y, z \in L$:

(1) $x \rightarrow (y \rightarrow z) = y \rightarrow (x \rightarrow z)$

(2) $x \rightarrow x = I$

(3) $x \rightarrow y = y' \rightarrow x'$

(4) $x \rightarrow y = y \rightarrow x = I$ implies $x = y$

(5) $(x \rightarrow y) \rightarrow y = (y \rightarrow x) \rightarrow x$

(6) $(x \vee y) \rightarrow z = (x \rightarrow z) \wedge (y \rightarrow z)$

(7) $(x \wedge y) \rightarrow z = (x \rightarrow z) \vee (y \rightarrow z)$

then $(L, \wedge, \vee, ', \rightarrow, O, I)$ is called a lattice implication algebra (LIA).

## 3. Linguistic truth-valued concept lattice

Linguistic truth-valued concept lattice is the combination of classical concept lattice and linguistic truth-valued lattice implication algebra and its ideological core is constructing linguistic truth-valued relation between objects and attributes. It can be used to directly deal with incomparability linguistic information and is totally different from classical fuzzy concept lattice. In this section, we study the definitions and theorems of linguistic truth-valued concept lattice and give an example to illustrate it.

**Definition 3.1** [16] Denote $MT =\{$*True* (*Tr* for short), *False* (*Fa* for short)$\}$, which is called as the set of meta truth values. An LIA defined on the set of meta truth values is called a meta linguistic truth-valued LIA, where $Fa<Tr$, the operation $'$ is defined as $Tr'=Fa$ and $Fa'=Tr$, the operation $\rightarrow$ is defined as

$$\rightarrow : MT \times MT \rightarrow MT,$$
$$x \rightarrow y = x' \vee y.$$

Based on reference [16], we can obtain the following definition:

**Definition 3.2** Denote $AD =\{$*Slightly* (*Sl* for short), *Very* (*Ve*), *Absolutely* (*Ab*)$\}$, which is called as the set of linguistic modifiers. An LIA defined on the chain $Sl<Ve<Ab$ is called an LIA with modifiers if its implication is Lukasiewicz implication.

In the following, denote $L_6 = AD \times MT$.

Let $L_3 = \{a_1, a_2, a_3\}$, $L_2 = \{b_1, b_2\}$. We can define two Lukasiewicz LIAs on them, respectively, and still denote them as $L_3$, $L_2$.

$$L_3: a_1 < a_2 < a_3, \ a_i \rightarrow_{(L_3)} a_j = a_{(3-i+j)\wedge 3}, \ a_i'^{(L_3)} = a_i \rightarrow_{(L_3)} a_j.$$
$$L_2: b_1 < b_2, \ b_i \rightarrow_{(L_2)} b_j = b_{(2-i+j)\wedge 2}, \ b_1'^{(L_2)} = b_2.$$

We can construct a new LIA by using the product of them, $L_3 \times L_2$, whose Hasse diagram is shown as Figure 1.

For any $(a_i, b_j), (a_k, b_m) \in L_6$, $(a_i, b_j) \rightarrow (a_k, b_m) = (a_i \rightarrow_{L_3} a_k, b_j \rightarrow_{L_2} b_m)$, $(a_i, b_j)' = (a_i'^{(L_3)}, b_j'^{(L_2)})$.

Let $I = (a_3, b_2)$, $A = (a_2, b_2)$, $B = (a_1, b_2)$, $C = (a_3, b_1)$, $D = (a_2, b_1)$, $O = (a_1, b_1)$, Define the mapping $f$ as $f: AD \times MT \rightarrow L_3 \times L_2$, where $f(Ab,Tr) = I$, $f(Ve,Tr) = A$, $f(Sl,Tr) = B$, $f(Sl,Fa) = C$, $f(Ve,Fa) = D$, $f(Ab,Fa) = O$. Then $f$ is a bijection. Denote its inverse mapping as $f^{-1}$. Define

$$x \vee y = f^{-1}(f(x) \vee f(y)), \ x \wedge y = f^{-1}(f(x) \wedge f(y)),$$
$$x \rightarrow y = f^{-1}(f(x) \rightarrow f(y)), \ x' = f^{-1}(f(x)').$$

It can be proved routinely that $(L_6, \vee, \wedge, ', \rightarrow)$ (denoted as $L_6$) is an *LIA*, and it is isomorphic to $L_3 \times L_2$, where $f$ is an isomorphic mapping from $L_6$ to $L_3 \times L_2$.

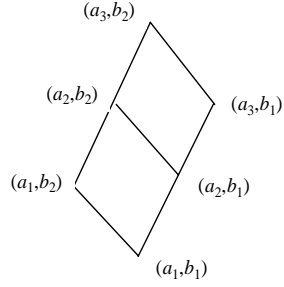

Figure 1. Hasse diagram of $L_3 \times L_2$.

**Definition 3.3** [16] The lattice implication algebra $L_6$ defined above is called a linguistic truth-valued *LIA* generated by *AD* and *MT*, denoted as *L-LIA*.

**Definition 3.4** A four-tuple $K = (G, M, L_6, \tilde{I})$ is called an n-ary linguistic truth-valued context, where $G = \{g_1, g_2, \cdots, g_r\}$ is the set of objects, $M = \{m_1, m_2, \cdots, m_s\}$ is the set of attributes, $L_6$ is an 6-ary linguistic truth-valued lattice implication algebra, $\tilde{I}$ is a fuzzy relation between $G$ and $M$, i.e., $\tilde{I} : G \times M \to L_6$.

Let $G$ be a non-empty objects set and $(L_6, \vee, \wedge, ', \to)$ an 6-ary linguistic truth-valued lattice implication algebra. Denote the set of all the fuzzy subsets on $G$ as $L_6^G$, for any $\tilde{A}_1, \tilde{A}_2 \in L_6^G$, $\tilde{A}_1 \subseteq \tilde{A}_2 \Leftrightarrow \tilde{A}_1(g) \leq \tilde{A}_2(g)$, $g \in G$, then $(L_6^G, \subseteq)$ is a partial ordered set.

Let $M$ be a non-empty attributes set and $(L_6, \vee, \wedge, ', \to)$ an 6-ary linguistic truth-valued lattice implication algebra. Denote the set of all the fuzzy subsets on $M$ as $L_6^M$, for any $\tilde{B}_1, \tilde{B}_2 \in L_6^M$, $\tilde{B}_1 \subseteq \tilde{B}_2 \Leftrightarrow \tilde{B}_1(m) \leq \tilde{B}_2(m)$, $m \in M$, then $(L_6^M, \subseteq)$ is a partial ordered set.

**Theorem 3.1** Let $K = (G, M, L_6, \tilde{I})$ be an 6-ary linguistic truth-valued context, $L_6$ be an 6-ary linguistic truth-valued lattice implication algebra, define mappings $f_1, f_2$ between $L_6^G$ and $L_6^M$,

$$\begin{cases} f_1 : L_6^G \to L_6^M, f_1(\tilde{A})(m) = \bigwedge_{g \in G}(\tilde{A}(g) \to \tilde{I}(g,m)) \\ f_2 : L_6^M \to L_6^G, f_2(\tilde{B})(g) = \bigwedge_{m \in M}(\tilde{B}(m) \to \tilde{I}(g,m)) \end{cases},$$

then for any $\tilde{A} \in L^G, \tilde{B} \in L^M$, $(f_1, f_2)$ is a Galois connection based on the 6-ary linguistic truth-valued lattice implication algebra.

**Definition 3.5** Let $K = (G, M, L_6, \tilde{I})$ be an 6-ary linguistic truth-valued context, denote the set $L(G, M, L_6, \tilde{I}) = \{(\tilde{A}, \tilde{B}) | f_1(\tilde{A}) = \tilde{B}, f_2(\tilde{B}) = \tilde{A}\}$, define $(\tilde{A}_1, \tilde{B}_1) \leq (\tilde{A}_2, \tilde{B}_2) \Leftrightarrow \tilde{A}_1 \subseteq \tilde{A}_2 (or \ \tilde{B}_2 \subseteq \tilde{B}_1)$.

**Theorem 3.2** Let $K = (G, M, L_6, \tilde{I})$ be an 6-ary linguistic truth-valued context, define the operations $\wedge$ and $\vee$ on $L(G, M, L_6, \tilde{I})$ as:

$$\bigwedge_i (\tilde{A}_i, \tilde{B}_i) = (\bigcap_i \tilde{A}_i, f_1 f_2(\bigcup_i \tilde{B}_i));$$

$$\bigvee_i (\tilde{A}_i, \tilde{B}_i) = (f_2 f_1(\bigcup_i \tilde{A}_i), \bigcap_i \tilde{B}_i),$$

then $L(G,M,L_n,\tilde{I})$ is a complete lattice.

**Example 3.1** A linguistic truth-valued context and its concept lattice based on the 6-ary linguistic truth-valued lattice implication algebra are shown as follows:

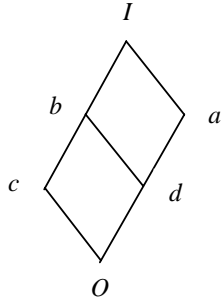

**Fig. 1.** Hasse Diagram of 6-ary lattice implication algebra $(L, \wedge, \vee, ', \rightarrow, O, I)$

**Table 1.** Implication operator of $L_6 = \{O, a, b, c, I\}$

| $\rightarrow$ | $O$ | $a$ | $b$ | $c$ | $I$ |
|---|---|---|---|---|---|
| $O$ | $I$ | $I$ | $I$ | $I$ | $I$ |
| $a$ | $c$ | $I$ | $I$ | $I$ | $I$ |
| $b$ | $b$ | $b$ | $I$ | $I$ | $I$ |
| $c$ | $a$ | $b$ | $b$ | $I$ | $I$ |
| $I$ | $O$ | $a$ | $b$ | $c$ | $I$ |

**Table 2.** Linguistic truth-valued context $K = (G, M, L_6, \tilde{I})$

|  | $m_1$ | $m_2$ | $m_3$ |
|---|---|---|---|
| $g_1$ | $a$ | $b$ | $I$ |
| $g_2$ | $b$ | $O$ | $a$ |

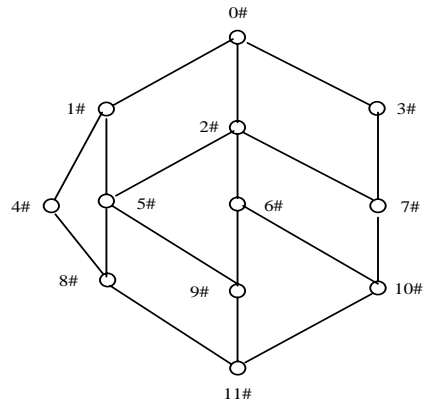

**Fig. 2.** Hasse Diagram of 6-ary linguistic truth-valued concept lattice of $K$

In this Hasse diagram, the linguistic truth-valued concepts are shown as follows:

0# $(\{I,I\},\{O,O,a\})$   1# $(\{a,I\},\{b,O,a\})$   2# $(\{I,a\},\{O,b,I\})$   3# $(\{I,b\},\{a,O,a\})$
4# $(\{a,a\},\{b,b,I\})$   5# $(\{a,b\},\{I,O,a\})$   6# $(\{b,b\},\{a,a,a\})$   7# $(\{I,O\},\{a,b,I\})$
8# $(\{a,O\},\{I,b,I\})$   9# $(\{O,b\},\{I,a,a\})$   10# $(\{b,O\},\{a,I,I\})$   11# $(\{O,O\},\{I,I,I\})$.

## 4. Tacit knowledge mining algorithm based on linguistic truth-valued concept lattice

In the pioneering work on the methods of mining tacit knowledge, many researchers mainly put emphasis on interview and observation, and in the whole process of tacit knowledge capture, accuracy and timeliness can not be guaranteed. Up to now, few studies on establishing a mathematical model for tacit knowledge. Taking into account the importance of tacit knowledge. in this section, we establish a concrete mathematical model and research the relevant theories.

*4.1 Tacit knowledge mining theory of linguistic truth-valued concept lattice*

**Definition 4.1** Let $K = (G, M, L_6, \tilde{I})$ be a 6-ary linguistic truth-valued context, the four-tuple $K_M = (G, M_+, L_6, \tilde{I}_+)$ is called a 6-ary linguistic truth-valued attribute extended context of $K$, where $G$ is a set of objects, $M_+ \supseteq M$ is an extended set of attributes, $L_6$ is a 6-ary linguistic truth-valued lattice implication algebra, $\tilde{I}_+$ is the fuzzy relation of $G$ and $M_+$, i.e., $\tilde{I}_+ : G \times M_+ \to L_6$ and $\forall g \in G$, $m \in M \cap M_+$, satisfying $g\tilde{I}m = g\tilde{I}_+ m$.

For the above linguistic truth-valued contexts, there relevant linguistic truth-valued concept lattices can be denoted by $L(G, M, L_6, \tilde{I})$, $L(G, M_+, L_6, \tilde{I})$, respectively. And denote the set:
$\tilde{\boldsymbol{A}} = \{\tilde{A} | (\tilde{A}, \tilde{B}) \in L(G, M, L_6, \tilde{I})\}$, $\tilde{\boldsymbol{B}} = \{\tilde{B} | (\tilde{A}, \tilde{B}) \in L(G, M, L_6, \tilde{I})\}$ ;
$\tilde{\boldsymbol{A}}_M = \{\tilde{A} | (\tilde{A}, \tilde{B}_+) \in L(G, M_+, L_6, \tilde{I})\}$, $\tilde{\boldsymbol{B}}_M = \{\tilde{B}_+ | (\tilde{A}, \tilde{B}_+) \in L(G, M_+, L_6, \tilde{I})\}$.

**Definition 4.2** Let $K = (G, M, L_6, \tilde{I})$ be a 6-ary linguistic truth-valued context, $K_M = (G, M_+, L_6, \tilde{I}_+)$ an 6-ary linguistic truth-valued attribute extended context. $K_M$ is called congener context of $K$ if $\tilde{\boldsymbol{A}}_M = \tilde{\boldsymbol{A}}$, accordingly, $L(K_M)$ is called congener linguistic truth-valued concept lattice of $L(K)$.

For a 6-ary linguistic truth-valued attribute extended context $K_M = (G, M_+, L_6, \tilde{I}_+)$, $(f_1, f_2)$ is a Galois connection between $G$ and $M_+$, $\forall \tilde{A} \in L_6^G$, $m \in M$ and $m_+ \in M_+$, denote:
$$f_1(\tilde{A})(m) = \tilde{B} \in L_6^M ;$$
$$f_1(\tilde{A})(m_+) = \tilde{B}_+ \in L_6^{M_+} .$$

**Theorem 4.1** Let $K_M = (G, M_+, L_6, \tilde{I}_+)$ be a 6-ary linguistic truth-valued attribute extended context of $K$, $\forall \tilde{A} \in L_6^G$, $m \in M$, $m_+ \in M_+$, $g \in G$, then $K_M$ is the congener context of $K$ if and only if $f_2(\tilde{B}_+)(g) = f_2(\tilde{B})(g)$.

**Proof.** $\forall \tilde{A} \in \tilde{\boldsymbol{A}}_M$, $\exists (\tilde{A}, \tilde{B}_+) \in L(G, M_+, L_6, \tilde{I})$, i.e., $(f_2(\tilde{B}_+)(g), \tilde{B}_+) \in L(G, M_+, L_6, \tilde{I})$; $\forall \tilde{A} \in \tilde{\boldsymbol{A}}$, $\exists (\tilde{A}, \tilde{B}) \in L(G, M, L_6, \tilde{I})$, i.e., $(f_2(\tilde{B})(g), \tilde{B}) \in L(G, M, L_6, \tilde{I})$, by definition 4.2, $K_M$ is the congener context of $K \Leftrightarrow \tilde{\boldsymbol{A}}_M = \tilde{\boldsymbol{A}} \Leftrightarrow f_2(\tilde{B}_+)(g) = f_2(\tilde{B})(g)$.

**Theorem 4.2** Let $K_M = (G, M_+, L_6, \tilde{I}_+)$ be a 6-ary linguistic truth-valued attribute extended context of $K$, $\forall \tilde{A} \in L_6^G$, $n \in M_+/M$, $g \in G$, then $K_M$ is the congener context of $K$ if and only if $f_2(\tilde{B})(g) \leq f_2(f_1(\tilde{A})(n))(g)$.

**Proof.** By theorem 4.1, $K_M$ is the congener context of $K$

$\Leftrightarrow f_2(\tilde{B}_+)(g) = f_2(\tilde{B})(g)$

$\Leftrightarrow f_2(\tilde{B})(g) \wedge f_2(f_1(\tilde{A})(n))(g) = f_2(\tilde{B})(g)$

$\Leftrightarrow f_2(\tilde{B})(g) \leq f_2(f_1(\tilde{A})(n))(g)$.

**Corollary 4.1** Let $K_M = (G, M_+, L_6, \tilde{I}_+)$ be a 6-ary linguistic truth-valued attribute extended context of $K$, $\forall \tilde{A} \in L_6^G$, $n \in M_+/M$, $g \in G$, then $K_M$ is the congener context of $K$, if $\exists m_j \in M$, $1 \leq j \leq s$, s.t., $f_2(f_1(\tilde{A})(m_j))(g) \leq f_2(f_1(\tilde{A})(n))(g)$.

**Theorem 4.3** Let $K_M = (G, M_+, L_6, \tilde{I}_+)$ be a 6-ary linguistic truth-valued attribute extended context of $K$, $\forall \tilde{A} \in L_6^G$, $n \in M_+/M$, $g \in G$, then $K_M$ is the congener context of $K$ if $\exists m_{j_1}, m_{j_2} \in M$, $1 \leq j_1, j_2 \leq s$, s.t., $\tilde{I}_+(g, n) = \tilde{I}(g, m_{j_1}) \wedge \tilde{I}(g, m_{j_2})$.

**Proof.** By theorem 4.2,

$f_2(f_1(\tilde{A})(n))(g)$

$= \bigwedge_{n \in M_+/M} (f_1(\tilde{A})(n) \to \tilde{I}_+(g, n))$

$= \bigwedge_{n \in M_+/M} \left( \bigwedge_{g \in G} (\tilde{A}(g) \to \tilde{I}_+(g, n)) \to \tilde{I}_+(g, n) \right)$

$= \bigwedge_{g \in G} (\tilde{A}(g) \to \tilde{I}(g, m_{j_1}) \wedge \tilde{I}(g, m_{j_2})) \to (\tilde{I}(g, m_{j_1}) \wedge \tilde{I}(g, m_{j_2}))$

$= ((f_1(\tilde{A})(m_{j_1}) \wedge f_1(\tilde{A})(m_{j_2})) \to \tilde{I}(g, m_{j_1})) \wedge ((f_1(\tilde{A})(m_{j_1}) \wedge f_1(\tilde{A})(m_{j_2})) \to \tilde{I}(g, m_{j_2}))$

$\geq (f_1(\tilde{A})(m_{j_1}) \to \tilde{I}(g, m_{j_1})) \wedge (f_1(\tilde{A})(m_{j_2}) \to \tilde{I}(g, m_{j_2}))$

$\geq \bigwedge_{m \in M} (f_1(\tilde{A})(m) \to \tilde{I}(g, m))$

$= f_2(\tilde{B})(g)$, so $K_M$ is the congener fuzzy context of $K$.

**Corollary 4.2** Let $K_M = (G, M_+, L_6, \tilde{I}_+)$ be a 6-ary linguistic truth-valued attribute extended context of $K$, $\forall \tilde{A} \in L_6^G$, $n \in M_+/M$, $g \in G$, then $K_M$ is the congener context of $K$ if $\exists m_{j_1}, m_{j_2}, \cdots, m_{j_k} \in M$, $1 \leq j_1, j_2, \cdots, j_k \leq s$, s.t., $\tilde{I}_+(g, n) = \bigwedge_{p=1}^{k} \tilde{I}(g, m_{j_p})$.

**Theorem 4.4** Let $K_M = (G, M_+, L_6, \tilde{I}_+)$ be a 6-ary linguistic truth-valued attribute extended context of $K$, $\forall \tilde{A} \in L_6^G$, $n \in M_+/M$, $g \in G$, then $K_M$ is the congener context of $K$ if $\tilde{I}_+(g,n) = I$.

**Proof.** By theorem 4.2,

$$f_2\left(f_1(\tilde{A})(n)\right)(g)$$
$$= \bigwedge_{n \in M_+/M}\left(f_1(\tilde{A})(n) \to \tilde{I}_+(g,n)\right)$$
$$= \bigwedge_{n \in M_+/M}\left(f_1(\tilde{A})(n) \to I\right)$$
$$= I \geq f_2(\tilde{B})(g), \text{ so } K_M \text{ is the congener context of } K.$$

*4.2. Tacit knowledge mining algorithm of linguistic truth-valued concept lattice*

For general fuzzy concept lattice which is constructed on the interval [0,1], attribute effective increased will inevitably change the number of fuzzy concepts and the structure of fuzzy concept lattice. But, for 6-ary linguistic truth-valued concept lattice, attribute conditional increased will not change the number of fuzzy concepts and the structure of concept lattice. About tacit knowledge, we explain this through the following algorithms.

**(1) Generation algorithm of** $K_M = (G, M_+, L_6, \tilde{I}_+)$:

Input: the 6-ary linguistic truth-valued context $K = (G, M, L_6, \tilde{I})$, let $M_+ = M$

Output: the 6-ary linguistic truth attribute extended context $K_M = (G, M_+, L_6, \tilde{I}_+)$

**Begin**
  **while** ($(G, M, L_6, \tilde{I}) \neq \Phi$) do
    Calculate the attribute values $\tilde{I}(g, m_j)$ of each attribute $m_j \in M$
    **for** each $\tilde{I}(g, m_{j_1}), \tilde{I}(g, m_{j_2}) \in K$ **do**
      **if** $\tilde{I}(g, m_{j_1}) \wedge \tilde{I}(g, m_{j_2}) \neq \Phi$ **then**
        $\tilde{I}_+(g, n) := \tilde{I}(g, m_{j_1}) \wedge \tilde{I}(g, m_{j_2})$
        $M_+ := M \cup \{n\}$
      **else**
        $M_+ = M$
      **endif;**
    **for** each $\tilde{I}(g, m_{j_1}), \tilde{I}(g, m_{j_2}), \cdots, \tilde{I}(g, m_{j_k}) \in K$ **do**
      **if** $\bigwedge_{p=1}^{k} \tilde{I}(g, m_{j_p}) \neq \Phi$ **then**
        $\tilde{I}_+(g, n) := \bigwedge_{p=1}^{k} \tilde{I}(g, m_{j_p})$
        $M_+ := M \cup \{n\}$
      **else**
        $M_+ = M$

    **endif;**
   **for** $L_6$ **do**
    **if** $\exists I \in L_6$ **then**
     $\tilde{I}_+(g,n) := I$
     $M_+ := M \cup \{n\}$
    **else**
     $M_+ = M$
    **endif;**
   **endif**
  **endfor;**
 **endfor;**
**end**

### (2) Generation algorithm of linguistic truth-valued concepts

Input: the 6-ary linguistic truth-valued concept lattice $L(G, M, L_6, \tilde{I})$

Output: the 6-ary linguistic truth-valued attribute extended concept lattice $L(G, M_+, L_6, \tilde{I}_+)$

**Begin**
 **while** ( $(G, M, L_6, \tilde{I}) \neq \Phi$ ) **do**
  Calculate the linguistic truth-valued concepts $(\tilde{A}, \tilde{B})$ of $K = (G, M, L_6, \tilde{I})$
  **for** each $(\tilde{A}, \tilde{B}) \in L(G, M, L_6, \tilde{I})$,
   denote $\tilde{A} = (\tilde{A}(g_1), \tilde{A}(g_2), \cdots, \tilde{A}(g_r)) \in \tilde{\mathbf{A}}$, $\tilde{B} = (\tilde{B}(m_1), \tilde{B}(m_2), \cdots, \tilde{B}(m_s)) \in \tilde{\mathbf{A}}$ **do**
   **if** $\exists n \in M_+$, s.t., $\tilde{I}_+(g,n) = \tilde{I}(g, m_{j_1}) \wedge \tilde{I}(g, m_{j_2})$, $m_{j_1}, m_{j_2} \in M$ **then**
    $\tilde{B}_+ = (\tilde{B}(m_1), \cdots, \tilde{B}(m_s), \tilde{B}(m_{j_1}) \wedge \tilde{B}(m_{j_2}))$
    $(\tilde{A}, \tilde{B}_+) \in L(G, M_+, L_6, \tilde{I}_+)$
   **else**
    $(\tilde{A}, \tilde{B}_+) \notin L(G, M_+, L_6, \tilde{I}_+)$
   **endif;**
   **if** $\exists n \in M_+$, s.t., $\tilde{I}_+(g,n) = \bigwedge_{p=1}^{k} \tilde{I}(g, m_{j_p})$, $m_{j_1}, m_{j_2}, \cdots, m_{j_k} \in M$ **then**
    $\tilde{B}_+ = \left( \tilde{B}(m_1), \cdots, \tilde{B}(m_s), \bigwedge_{p=1}^{k} \tilde{I}(g, m_{j_p}) \right)$
    $(\tilde{A}, \tilde{B}_+) \in L(G, M_+, L_6, \tilde{I}_+)$
   **else**
    $(\tilde{A}, \tilde{B}_+) \notin L(G, M_+, L_6, \tilde{I}_+)$
   **endif;**
    **if** $\exists n \in M_+$, s.t., $\tilde{I}_+(g,n) = I$, $I \in L_n$ **then**
     $(\tilde{A}, \tilde{B}_+) \in L(G, M_+, L_6, \tilde{I}_+)$
    **else**

$$(\tilde{A}, \tilde{B}_+) \notin L(G, M_+, L_6, \tilde{I}_+)$$
              endif;
           endif
        endif;
     endfor;
  end

**Example 4.1** For Table 2, a 6-ary linguistic truth-valued context $K = (G, M, L_6, \tilde{I})$, we can compute its congener context as Table 3.

**Table 3.** The congener fuzzy context $K_M = (G, M_+, L_6, \tilde{I}_+)$ of $K$

|       | $m_1$ | $m_2$ | $m_3$ | $m_4$ | $m_5$ |
|-------|-------|-------|-------|-------|-------|
| $g_1$ | $a$   | $b$   | $I$   | $a$   | $I$   |
| $g_2$ | $b$   | $O$   | $a$   | $O$   | $I$   |

In this congener context $K_M$, $M_+ = \{m_1, m_2, \cdots, m_5\}$, $\tilde{I}_+ : G \times M_+ \to L_6$, and it follows that: $\forall g \in G$, $\tilde{I}_+(g, m_4) = \tilde{I}(g, m_1) \wedge \tilde{I}(g, m_2)$, $\tilde{I}_+(g, m_5) = I$. According to the theorems and algorithms of attribute increased, the linguistic truth-valued concepts of $K_M$ are easily derived from the relevant linguistic truth-valued concepts of $K$ as follows:

0# $(\{I,I\},\{O,O,a,O,I\})$     1# $(\{a,I\},\{b,O,a,O,I\})$     2# $(\{I,a\},\{O,b,I,O,I\})$
3# $(\{I,b\},\{a,O,a,O,I\})$     4# $(\{a,a\},\{b,b,I,b,I\})$     5# $(\{a,b\},\{I,O,a,O,I\})$
6# $(\{b,b\},\{a,a,a,a,I\})$     7# $(\{I,O\},\{a,b,I,O,I\})$     8# $(\{a,O\},\{I,b,I,b,I\})$
9# $(\{O,b\},\{I,a,a,a,I\})$   10# $(\{b,O\},\{a,I,I,a,I\})$   11# $(\{O,O\},\{I,I,I,I,I\})$.

The structure of linguistic truth-valued concept lattice established based on Table 3 is the same to that based on Table 2, so, comparing Table 3 and Table 2, we can easily capture tacit knowledge (attribute $m_4$ and attribute $m_5$) different from explicit knowledge existing in Table 2.

## 5. Conclusions

As the theoretical premise of effective mining tacit knowledge, our research about the model construction of 6-ary linguistic truth-valued concept lattice is helpful to provide a useful mathmatical tool. From the definitions of attribute extended context and congener context, this paper researches the necessary and sufficient conditions of generating tacit knowledge, and under this condition, we can obtain the invariable structure of linguistic truth-valued concept lattice. The algorithms of generating linguistic truth-valued context and of establishing linguistic truth-valued concept lattice based on these conditions are proposed.

**Acknowledgements**

This work is partially supported by the National Natural Science Foundation of P.R. China (Grant No. 71101049, 60875034, 61175055) .